\documentclass[10pt,twocolumn,letterpaper]{article}

\usepackage[pagenumbers]{cvpr} %
\makeatletter
\@namedef{ver@everyshi.sty}{}
\makeatother

\usepackage{graphicx}
\usepackage{amsmath}
\usepackage{amssymb}
\usepackage{booktabs}
\usepackage{mathrsfs}
\usepackage{algorithm}
\usepackage{algpseudocode}
\usepackage[utf8]{inputenc}
\usepackage{graphicx}
\usepackage{booktabs}
\usepackage{witharrows}
\usepackage{rotating}  
\usepackage{multirow}
\usepackage{pifont}
\usepackage{bbding}
\usepackage{adjustbox}
\usepackage{colortbl}
\usepackage{caption}
\usepackage{xcolor}
\usepackage{bm}
\definecolor{bb}{rgb}{0.0, 0.0, 0.5}

\definecolor{Gray}{gray}{0.9}

\def\etal{\emph{et al.}}

\definecolor{Gray}{gray}{0.9}

\newcommand{\chong}[1]{\textcolor{black}{#1}}
\newcommand{\lanqing}[1]{{\color{black}#1}}
\usepackage[pagebackref,breaklinks,colorlinks]{hyperref}

\usepackage[capitalize]{cleveref}
\crefname{section}{Sec.}{Secs.}
\Crefname{section}{Section}{Sections}
\Crefname{table}{Table}{Tables}
\crefname{table}{Tab.}{Tabs.}

\usepackage{color,soul}
\usepackage{xcolor}

\newcommand{\wh}[1]{\textcolor{black}{#1}}
\newcommand{\bihan}[1]{\textcolor{black}{#1}}

\def\cvprPaperID{2757} %
\def\confName{CVPR}
\def\confYear{2023}

\begin{document}

\title{ShadowDiffusion: When Degradation Prior Meets Diffusion Model for\\ Shadow Removal}
\author{Lanqing Guo\textsuperscript{\rm 1},
    Chong Wang\textsuperscript{\rm 1},
    Wenhan Yang\textsuperscript{\rm 2},
	Siyu Huang\textsuperscript{\rm 3},
	Yufei Wang\textsuperscript{\rm 1},
	  Hanspeter Pfister\textsuperscript{\rm 3},
  Bihan Wen\textsuperscript{\rm 1}\\
\textsuperscript{\rm 1}Nanyang Technological University, Singapore \\
 \textsuperscript{\rm 2}Peng Cheng Laboratory, China \quad \textsuperscript{\rm 3}Harvard University, USA\\
{\tt\small 	\{lanqing001, wang1711, yufei001, bihan.wen\}@ntu.edu.sg,}\\
{\tt\small 
    yangwh@pcl.ac.cn,
   huang@seas.harvard.edu, pfister@g.harvard.edu}
}
\maketitle

\begin{abstract}
   Recent deep learning methods have achieved promising results in image shadow removal.
However, \bihan{their restored images} \wh{still suffer from} unsatisfactory boundary artifacts, \wh{due to the lack of degradation prior \bihan{embedding} and the deficiency in modeling capacity}.
\bihan{Our work addresses these issues by proposing a unified diffusion framework that integrates both the image and degradation priors for highly effective shadow removal.}
\wh{In detail,} we first propose a shadow degradation model, which inspires us to build a novel unrolling diffusion model, dubbed \textit{ShandowDiffusion}. 
\wh{It remarkably improves the model's capacity in shadow removal via \wh{progressively} \wh{refining} the desired output with both degradation prior and diffusive generative prior, which by nature can serve as a new strong baseline for image restoration.}
\bihan{Furthermore, ShadowDiffusion progressively refines the estimated shadow mask as an auxiliary task of the diffusion generator, which leads to more accurate and robust shadow-free image generation.
}
We conduct extensive experiments on three popular public datasets, including ISTD, ISTD+, and SRD, to validate our method's effectiveness.
Compared to the state-of-the-art \wh{methods}, our model achieves a significant improvement in terms of PSNR, increasing from 31.69dB to 34.73dB over SRD dataset.
\end{abstract}

\section{Introduction}
\label{sec:intro}
\begin{figure}[!t]
\centering
\includegraphics[width=.95\linewidth]{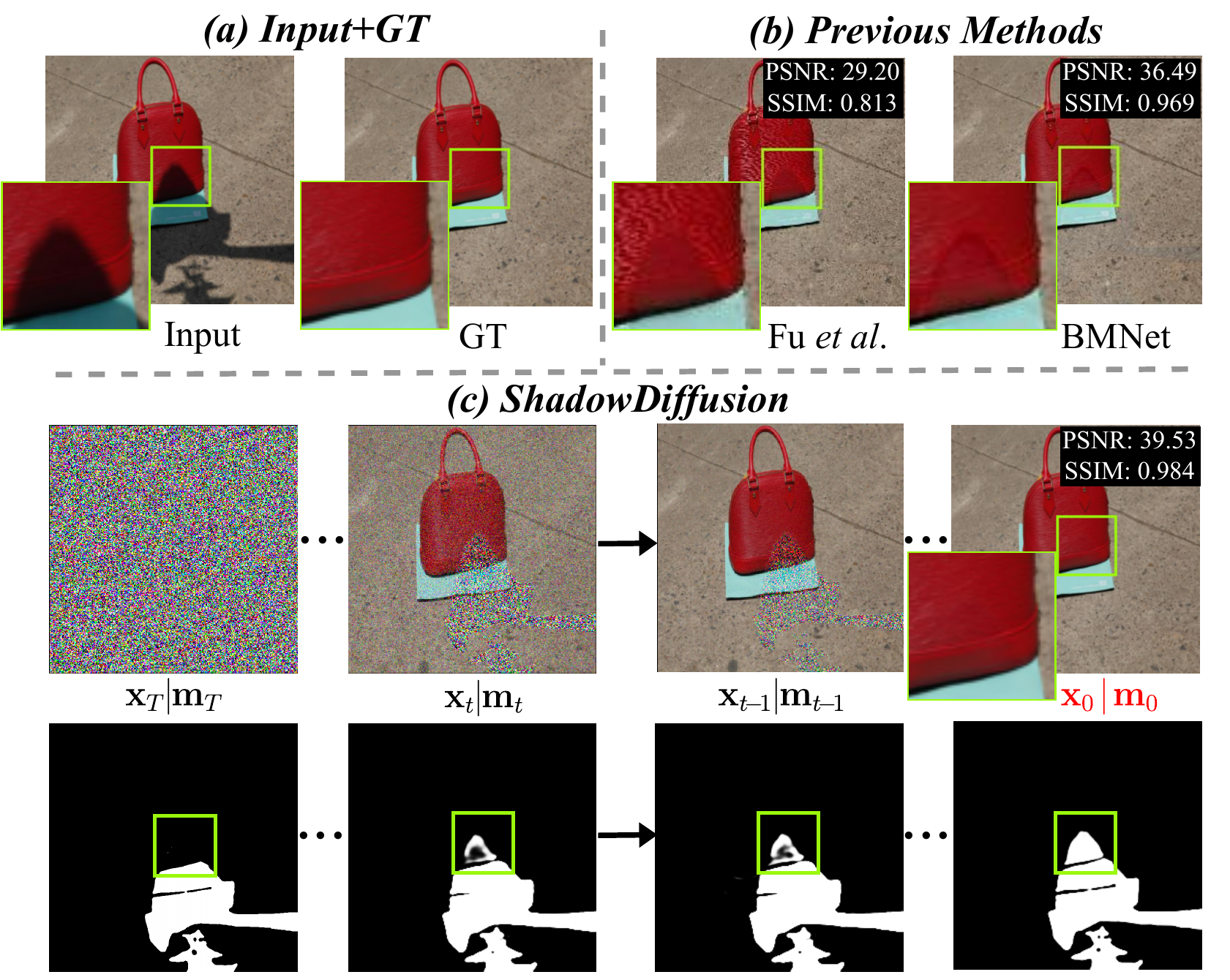} 
\vspace{-2mm}
\caption{(a) Input shadow image and corresponding ground truth shadow-free image, (b) shadow removal results of two most recent competing methods Fu~\etal~\cite{fu2021auto} and BMNet~\cite{zhu2022bijective}, and (c) our proposed ShadowDiffusion iteratively ($T\!\rightarrow \!0$) restores the shadow-free image and refines the shadow mask, in which the $\mathbf{x}_0$ and $\mathbf{m}_0$ are the final enhanced result and refined mask, respectively. 
 }%
\vspace{-0.2cm}
\label{fig:intro} 
\end{figure}

Shadow removal aims to enhance visibility of the image shadow regions, pursuing a consistent illumination distribution between shadow and non-shadow regions.
Deep learning-based methods~\cite{fu2021auto,chen2021canet,zhu2022bijective} achieved superior performance \wh{recently} by \wh{fully} utilizing the power of large collections of data.
\bihan{While most of the existing methods focused on learning the discriminative models for shadow removal, modeling the underlying distribution of nature images is overlooked in their restoration process.}
\bihan{Consequently, the shadow removal results usually contain severe boundary artifacts and remaining shadow patterns, as shown in Figure~\ref{fig:intro}(b).}
Though the adversarial loss can alleviate this issue, these approaches~\cite{wang2018stacked,hu2019mask} require careful adjustment during training, might overfit certain
visual features or data distribution, and might hallucinate new content and artifacts.
\bihan{Very recently, various diffusion models, such as the popular diffusion denoising diffusion probability model (DDPM)~\cite{ho2020denoising}, have gained wide interest in the field of low-level vision~\cite{saharia2022image,saharia2022palette}.
Comparing to other deep generative models, diffusion models are more powerful for modeling image pixel distribution, which provides great potential for significantly improving visual quality and benefits high-quality image restoration.
However, no work to-date has exploited diffusion models for shadow removal tasks.}

\bihan{Moreover, there are two major limitations in existing shadow removal methods: First, the shadow degradation prior that reflects its corresponding physical properties has not been well exploited in deep learning. Though recent work~\cite{le2019shadow} attempted to incorporate simple shadow model as a linear and uniform degradation, such an assumption is too restrictive for restoring real shadow images subjective to complicated lighting conditions.}
\bihan{Second, most of the deep shadow removal methods requires an estimated shadow mask as the inputs, which are either provided by the benchmark datasets~\cite{wang2018stacked} or generated by a pre-trained shadow detector~\cite{cun2020towards}. However, these mask estimates are usually inaccurate, \eg, wrong indicators near the boundary or small shadow objects.
Even the carefully hand-crafted masks sometimes contain coarse boundaries. 
Since existing methods blindly rely on the estimated masks without exploiting their correlation to the actual shadow images for refinement, there are usually severe boundary artifacts in their shadow removal results~\cite{fu2021auto,zhu2022bijective}, as shown in Figure~\ref{fig:intro}(b). 
}

To alleviate the challenges in shadow removal, we first introduce a general shadow model of spatially-variant degradation, by decomposing the degradation matrix into the
shadow mask and shadow intensities.
Based on the new shadow model, 
we propose a novel unrolling diffusion-based shadow removal framework, called ShadowDiffusion, which integrates both the generative and degradation priors.
Specifically, we formulate the shadow removal problem as to \textit{jointly} pursue the shadow-free image and refined shadow mask.
Mask refinement is designed as an auxiliary task of the diffusion generator to progressively refine the shadow mask along with shadow-free image restoration in an interactive manner as shown in Figure~\ref{fig:intro}(c).
After that, we further propose an unrolling-inspired diffusive sampling strategy to explicitly integrate the degradation prior into the diffusion framework.
Experimental results show that ShadowDiffusion can \wh{achieve  superior performance} consistently over the three widely-used shadow removal datasets and \wh{significantly outperform} the state-of-the-art methods.
Besides, our model can be applied to other image enhancement tasks, \eg, low-light image enhancement and exposure correction.
Our main contributions are summarized as follows:
\begin{itemize}
    \item We propose the first diffusion-based model for shadow removal. 
    A novel dynamic mask-aware diffusion model (DMDM) is introduced to jointly pursue a shadow-free image and refined shadow mask, which leads to robust shadow removal even with an inaccurate mask estimate. 
    
    \item We further propose an unrolling-inspired diffusive sampling strategy to explicitly integrate the shadow degradation prior into the intrinsic iterative process of DMDM.
    \item Extensive experimental results on the public ISTD, ISTD+, and SRD datasets show that the proposed ShadowDiffusion outperforms the state-of-the-art shadow removal methods by large margins. Besides, our method can be generalized to \wh{a series of} image enhancement tasks.
\end{itemize}

\section{Related Work}
\noindent\textbf{Shadow removal.}
Classic shadow removal methods usually employed various handcrafted priors, \eg, image gradients~\cite{gryka2015learning}, illumination~\cite{zhang2015shadow}, and regions~\cite{guo2012paired}, for enhancing the illumination of shadow regions.  
Those methods \wh{are} built under an ideal assumption, leading to obvious shadow boundary artifacts when transferring to real-world cases.

Taking advantage of the powerful ability in learning mappings from the training pairs, deep learning-based methods~\cite{le2019shadow,ding2019argan,hu2019mask} achieved superior performance in recent years.
One group of works still reconstructed the shadow-free image under a physical illumination model with global degradation.
For instance, 
Le~\etal~\cite{le2019shadow} ideally applied the physical linear transformation model to enhance the shadow region and reconstructed the shadow-free image by image decomposition.
Fu \etal~\cite{fu2021auto} proposed an over-exposure fusion way for shadow removal, where the proposed model can smartly blend a series of over-enhanced shadow images as well as the original shadow image by a learnable pixel-wise weighting map.
However, such a global degradation model is too strict since real-world illumination degradation is always non-uniform. 
Besides, generative adversarial network techniques~\cite{wang2018stacked,hu2019mask,liu2021from} are applied to enhance the reality of enhanced results. 
However, the results of these methods always suffered from color distortions and might hallucinate new content and artifacts.

\begin{figure*}[!t]
\centering
\vspace{-3mm}
\includegraphics[width=.95\linewidth]{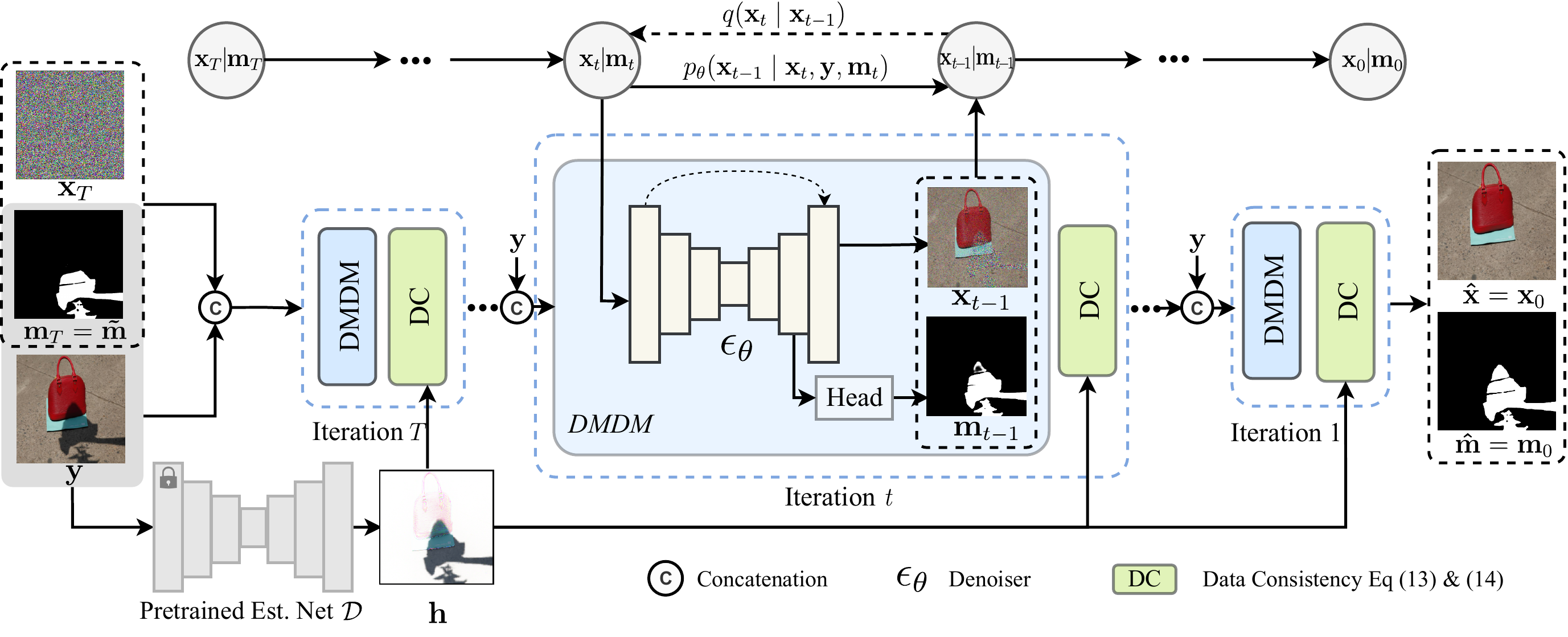} 
\vspace{-2mm}
\caption{
\lanqing{
Illustration of the proposed ShadowDiffusion, in which the training (dashed line) and sampling (solid line) processes \wh{are} detailed in Algorithm~\ref{algo_train} and  Algorithm~\ref{algo}, respectively.
Each sampling iteration consists of the sampling of dynamic mask-aware diffusion model (DMDM) and data consistency (DC) steps (Here DC corresponds to Eq (\ref{eq-solver_sub1}) \& (\ref{eq-solver_sub2})).}
}
\vspace{-0.2cm}
\label{fig:framework} 
\end{figure*}

\noindent\textbf{Diffusion model for image restoration.}
Diffusion-based generative models~\cite{sohl2015deep} recently produced amazing results with improvements adopted in denoising diffusion probabilistic models~\cite{ho2020denoising}, which becomes increasingly influential in the field of 
low-level vision tasks, such as super-resolution~\cite{kawar2022denoising,saharia2022image}, inpainting~\cite{lugmayr2022repaint}, and colorization~\cite{saharia2022palette}. 
Saharia~\etal~\cite{saharia2022image} introduced \wh{a} denoising diffusion probabilistic model to image super-resolution and achieved \wh{better} performance compared with the state-of-the-art Generative Adversarial Network (GAN)~\cite{goodfellow2020generative} based methods.
Inspired by conditional generation models~\cite{mirza2014conditional,lugmayr2020srflow},
Pallette~\cite{saharia2022palette} was proposed as a general image-to-image framework to solve the image restoration with conditional denoising diffusion probability models.
Ozan~\etal~\cite{ozdenizci2022restoring} presented a patch-based diffusion model for weather removal \wh{that} enables the size-agnostic processing, which employed a guided denoising process across
overlapping patches during inference.

However, most of these methods focus on synthetic degradation, such as image colorization, image inpainting, and super-resolution, in which it is very easy to simulate large-scale training pairs on existing natural image \wh{datasets} to train diffusion-based models.
In this paper, we explore the real-world shadow removal problem with limited training pairs. A novel dynamic mask-aware diffusion model with \wh{the} stricter and iteratively refined \wh{conditions} is proposed to address \wh{the} above problem.

\noindent\textbf{Deep unrolling methods.}
Consistency of the predictions with respect to the degradation model is crucial for reliably solving ill-posed restoration tasks. 
Deep unrolling, by incorporating the known degradation model into the deep networks via \wh{an} iterative optimization algorithm, has demonstrated remarkable performance on various inverse problems. 
For example, Karol~\etal~\cite{lista} proposed to unroll the iterative shrinkage thresholding algorithm (ISTA) for sparse coding, which demonstrated promising results on super-resolution~\cite{liu2016robust}.
Yang~\etal~\cite{admmnet} introduced an unrolling network describing the data flow graphs in the iterative procedures of Alternating Direction Method of Multipliers (ADMM) for magnetic resonance imaging (MRI) reconstruction. 
Based on half-quadratic splitting, Zhang~\etal~\cite{zhang2020deep} proposed an unfolding scheme that enables a single network to address different scale factors in \wh{the} super-resolution task.
Compared with model-free learning based methods, \wh{the} deep unrolling scheme integrates the degradation constraint into the learning model by iteratively regularizing the network output according to the model prior.

\section{ShadowDiffusion}
\bihan{We present the proposed ShadowDiffusion, by first introducing our shadow degradation model. 
Then, the dynamic mask-aware diffusion model (DMDM) and its training process are presented, which can \wh{predict} the shadow-free image \wh{jointly} with progressive mask refinement.
Finally, we introduce the unrolling-inspired diffusive sampling based on the DMDM which integrates the diffusive generative model and shadow degradation prior.
}

\subsection{Shadow Degradation Model}
A shadow region of an image $\mathbf{y}$ is caused by partial or complete \wh{occlusions}.
Inspired by Retinex theory~\cite{land1977retinex}, 
classic methods adopted a simple shadow degradation,
in which shadow images $\mathbf{y}$ \wh{is} formed by applying an illumination change surface $\mathbf{a}$ to the shadow-free image $\mathbf{x}$ as follows:
\begin{small}
\begin{align}
     \mathbf{y} = \mathbf{a} \cdot \mathbf{x}  \;,
      \label{eq:problem}
\end{align}
\end{small}
where $\cdot$ denotes the element-wise multiplication. 
Here $\mathbf{a}$ is strictly assumed to be \wh{$1$} in the lit
area, and a constant $a \in(0,1)$ in the umbra area.
(\ref{eq:problem}) is usually too restrictive as the natural lighting are mostly non-uniform in practice.

 \wh{In general, illumination} degradation should be spatially-variant and highly dependent \wh{on} shadow mask information.
\bihan{Thus, we propose a new shadow degradation model as}
\begin{small}
\begin{align}
     \mathbf{y}= \mathbf{h} \cdot \mathbf{x} = \mathbf{w} \cdot \mathbf{m}\cdot \mathbf{x} +(\mathbf{1}-\mathbf{m})\cdot \mathbf{x}\;.
      \label{eq:shadow_model}
\end{align}
\end{small}
Here, $\mathbf{h}$ denotes the pixel-wise illumination degradation map, which can be decomposed into the shadow mask $\mathbf{m}$ and illumination weight $\mathbf{w}$. The shadow mask $\mathbf{m}$ indicates the shadow locations that shadow regions are $1$ and the rest are $0$.
Our Model \eqref{eq:shadow_model} \wh{owns the following advantages}:
\begin{itemize}
    \item The shadow image $\mathbf{y}$ can be modeled as \wh{a} non-uniform illumination transformation on shadow-free image $\mathbf{x}$ under the pixel-wise degradation map $\mathbf{h}$.
    \wh{This degradation prior provides richer information than the uniform degradation for shadow removal.}
    \item The shadow mask $\mathbf{m}$ provides shadow location information, which have \wh{a direct and critical effect on} the shadow degradation $\mathbf{h}$, \wh{which therefore significantly affect the estimation of the desired shadow-free image $\mathbf{\hat{x}}$.}
    \wh{As a wrong shadow mask leads to an inaccurate degradation map, which is quite common in practice, our degradation model flexibly enables to embed the mask refinement as an auxiliary task, and make the refinement and shadow-free image restoration mutually beneficial.}

\end{itemize}

\vspace{-10pt}
\subsection{Dynamic Mask-Aware Diffusion Model}
\label{sec:dmdm}
\vspace{-2pt}
\bihan{The shadow mask is \wh{crucial} for shadow removal.
\wh{It indicates} the exact location of shadow regions according to \eqref{eq:shadow_model}.
In other words, inaccurate shadow mask inputs will directly affect the shadow removal outcomes.}
Thus, we remodel the shadow removal as a \wh{joint task} to pursue a shadow-free image and refined mask, in which mask refinement would be an auxiliary task of the diffusion generator to progressively refine the shadow mask along with shadow-free image generation.
Different from previous conditional diffusion-based image restoration works~\cite{saharia2022image,saharia2022palette} generating the underlying image with an invariable condition, we propose a dynamic mask-aware diffusion model (DMDM) to \wh{progressively} generate the shadow-free image and refined conditions (masks).

We first revisit the previous conditional diffusion model~\cite{saharia2022image}, which learns a conditional reverse process $p_\theta(\mathbf{x}_{0:T}|{\mathbf{y}})$ without modifying the diffusion process $q(\mathbf{x}_{1:T}|\mathbf{x}_{0})$ for $\mathbf{x}$, such that the sampled image has high fidelity to the data distribution conditioned on $\mathbf{y}$.
During training\wh{,} we sample $(\mathbf{x}_0, \mathbf{y}, \mathbf{\tilde{m}}) \sim q(\mathbf{x}, \mathbf{y}, \mathbf{\tilde{m}})$ from a triplet data distribution (\eg, a shadow-free image $\mathbf{x}$, shadow image $\mathbf{y}$, and corresponding initial shadow mask $\mathbf{\tilde{m}}$).
Our training approach is outlined in Algorithm~\ref{algo_train}, in which we learn the dynamic mask-aware reverse process:
\begin{small}
\begin{align}
\setlength\abovedisplayskip{2pt}%
\setlength\belowdisplayskip{2pt}
p_\theta\left(\mathbf{x}_{0: T} \!\mid \!\mathbf{y}, \mathbf{m}_{0:T}\right)=p\left(\mathbf{x}_T\right) \prod_{t=1}^T p_\theta\left(\mathbf{x}_{t-1} \!\mid\! \mathbf{x}_t, \mathbf{y}, \mathbf{m}_t\right)\;.
\end{align}
\end{small}
We can marginalize the Gaussian diffusion process to sample intermediate $\mathbf{x}_t$ terms directly from shadow-free image $\mathbf{x}_0$ through $\mathbf{x}_t = \sqrt{\bar{\alpha}_t} \mathbf{x}_0+\sqrt{1-\bar{\alpha}_t} \bm{\epsilon}$, where $\beta_t$ is the noise schedule, $\alpha_t = 1-\beta_t$, $\bar{\alpha}_t=\prod_{i=1}^t\alpha_i$, and $\bm{\epsilon} \sim \mathcal{N}(\mathbf{0},\mathbf{I})$ has the same dimensionality as $\mathbf{x}_0$.
The denoiser $\bm{\epsilon}_\theta$ takes the shadow image $\mathbf{y}$, the intermediate variable $\mathbf{x}_t$, and the time step $t$ as input to predict the noise map $\mathbf{e}_t$ and the refined mask $\mathbf{m}_t$ as follows:
\begin{small}
\begin{align}
\setlength\abovedisplayskip{2pt}%
\setlength\belowdisplayskip{2pt}
    \mathbf{e}_t, \mathbf{m}_t = \bm{\epsilon}_\theta\left(\sqrt{\bar{\alpha}_t} \mathbf{x}_0+\sqrt{1-\bar{\alpha}_t} \bm{\epsilon},  \mathbf{y}, \mathbf{\tilde{m}}, t\right)\;.
\end{align}
\end{small}
\wh{As} the shadow mask information is highly dependent \wh{on the shadow-free image generation, we build a model to perform the shadow-free image prediction and mask refinement jointly.}
We add a mask prediction head after the last layer of $\bm{\epsilon}_\theta$, with one $1\times 1$ convolution layer and one Sigmoid function to predict the refined mask.
Following~\cite{ho2020denoising}, the diffusive objective function is
\begin{small}
\begin{align}
    \mathcal{L}_{diff}=\mathbb{E}_{\mathbf{x}_0, t, \bm{\epsilon}}\left\|\mathbf{e}_t - \bm{\epsilon}\right\|^2_F.
\end{align}
\end{small}
Besides, according to the shadow and shadow-free image pairs in the training stage, we can adopt a ground truth shadow mask as \wh{a} reference \wh{to constrain the rationality of the refined mask}
\begin{small}
\begin{align}
    \mathcal{L}_{mask}=\mathbb{E}_{t \sim [1,T]}\left\|\mathbf{m}_t - \mathbf{m}_{gt}\right\|^2_F,
\end{align}
\end{small}
where the ground truth shadow mask $\mathbf{m}_{gt}$ can be obtained by binarizing the residual map between shadow and shadow-free images $\mathbf{m}_{gt}=\begin{cases}
 1&  \mathbf{x}- \mathbf{y}>0.1\;, \\ 
 0& \text{otherwise}\;.
\end{cases}$

\wh{The hybrid objective function $\mathcal{L}_{total}$ is obtained by combining the above losses, which guides the training of the denoiser $\bm{\epsilon}_\theta$ in our DMDM as follows,}
\begin{small}
\begin{align}
    \mathcal{L}_{total}= \mathcal{L}_{diff} + \lambda \mathcal{L}_{mask} \;,
\end{align}
\end{small}
where $\lambda$ are the weighting \wh{coefficient} to balance the influence of each term.

\begin{algorithm}[t]
\hspace*{\algorithmicindent}\noindent \textbf{Input:} shadow image $\mathbf{y}$, shadow-free image $\mathbf{x}$, and initial mask $\mathbf{\tilde{m}}$.
\caption{Dynamic mask-aware diffusion training.}\label{algo_train}
\begin{algorithmic}[1]
        \While{not converged}
        \State $t \sim \text{Uniform}\{1, \ldots, T\}$
        \State $\bm{\epsilon} \sim \mathcal{N}(\mathbf{0},\mathbf{I})$
        \State $\mathbf{e}_t, \mathbf{m}_t = \bm{\epsilon}_\theta\left(\sqrt{\bar{\alpha}_t} \mathbf{x}_0+\sqrt{1-\bar{\alpha}_t} \bm{\epsilon},  \mathbf{y}, \mathbf{\tilde{m}}, t\right)$
        \State Perform Gradient descent steps on $\nabla_\theta \;\mathcal{L}_{total}(\theta)$
        \EndWhile
      \State $\textbf{return}\;\; \theta$
    
\end{algorithmic}
\end{algorithm}
\begin{algorithm}[!t]
\hspace*{\algorithmicindent}\noindent \textbf{Input:} shadow image $\mathbf{y}$, initial mask $\mathbf{\tilde{m}}$, diffusion model $\bm{\epsilon}_\theta$, number of implicit sampling iterations $T$, $\mathbf{z}_T \sim \mathcal{N}(\mathbf{0}, \mathbf{I})$, $\mathbf{v}_T=\mathbf{\tilde{m}}$, and initial parameters $\psi$, $\phi$, and $\rho$.
\caption{Unrolling-inspired diffusive sampling.}\label{algo}
\begin{algorithmic}[1]
      \For{$t =T, \ldots, 1$}
        \State $\mathbf{e}_{t-1}, \mathbf{m}_{t-1} = \bm{\epsilon}_\theta\left(\mathbf{z}_{t}, \mathbf{y}, \mathbf{v}_{t}, t\right)$
        \State $\mathbf{x}_{t-1} \!\!= \!\!\sqrt{\bar{\alpha}_{t-{1}}}\left(\frac{\mathbf{z}_{t}-\sqrt{1-\bar{\alpha}_t} \cdot \mathbf{e}_{t-1}}{\sqrt{\bar{\alpha}_t}}\right)
       \! +\!\sqrt{1-\bar{\alpha}_{t-{1}}} \cdot \mathbf{e}_{t-1}$
        \State update $\mathbf{z}_{t-1}$ with Eq~(\ref{eq-solver_sub1}).
        \State update $\mathbf{v}_{t-1}$ with Eq~(\ref{eq-solver_sub2}).
      \EndFor
      \State $\textbf{return}\;\; \mathbf{x}_t, \  \mathbf{m}_t$
\end{algorithmic}
\end{algorithm}

\subsection{Unrolling-Inspired Diffusive Sampling}
Based on the shadow degradation model~(\ref{eq:shadow_model}), we formulate the shadow removal as \wh{a degradation prior guided} model, where \wh{regularization terms} are inferred
by \wh{a} learnable conditional generative diffusion model instead of using hand-crafted priors under Maximum A Posteriori (MAP) framework.
 By considering the provided initial mask $\mathbf{\tilde{m}}$ may be coarse or \wh{inaccurate}, the shadow mask $\mathbf{m}$ would be \wh{iteratively} refined along with desired shadow-free image $\mathbf{x}$ optimization, which can be obtained by minimizing the following energy function with joint image-mask regularizer:
 \begin{small}
\begin{align}
\setlength\abovedisplayskip{3pt}%
\setlength\belowdisplayskip{3pt}
   \nonumber\min_{\mathbf{x}, \mathbf{m}, \mathbf{z}, \mathbf{v}}{\frac{1}{2}\lVert{\mathbf{h} \cdot \mathbf{z}-\mathbf{y}}\rVert^2_F}+&\psi\mathcal{R}{([\mathbf{x}|\mathbf{m}])} + \chong{\frac{\phi}{2}}\lVert{\mathbf{v} - \mathbf{\tilde{m}}}\rVert^2_F \\ &\quad \text{s.t.} \;  \mathbf{x}=\mathbf{z}, \;\mathbf{m}=\mathbf{v}\;, \label{eq-opt_problem}
\end{align}
\end{small}
where $\mathbf{z}$ and $\mathbf{v}$ are the auxiliary variables that convert~(\ref{eq-opt_problem}) into a constrained problem and $\mathcal{R}(\cdot)$ is the regularizer capturing assumed joint image and mask priors. $[\cdot|\cdot]$ denotes the concatenation operation. $\psi$ and $\phi$ are trade-off parameters. Here we assume that the degradation matrix $\mathbf{h}$ can be estimated by a pre-trained degradation estimation network as $\mathbf{h} =\mathcal{D}(\mathbf{y}, \mathbf{\tilde{m}}) $.
To deal with the equality constraints, two quadratic penalty
terms are introduced, and \wh{the} problem is rewritten as follows:
\begin{small}
\begin{align}
\setlength\abovedisplayskip{3pt}%
\setlength\belowdisplayskip{3pt}
   \nonumber\min_{\mathbf{x}, \mathbf{m}, \mathbf{z}, \mathbf{v}}&{\frac{1}{2}\lVert{\mathbf{h} \cdot \mathbf{z}-\mathbf{y}}\rVert^2_F}+\psi\mathcal{R}{([\mathbf{x}|\mathbf{m}])} + \chong{\frac{\phi}{2}}\lVert{\chong{\mathbf{v}} - \mathbf{\tilde{m}}}\rVert^2_F \\ &\quad + \chong{\frac{\rho_1}{2}}\lVert{\mathbf{x} - \mathbf{z}}\rVert^2_F+ \chong{\frac{\rho_2}{2}}\lVert{\mathbf{m} - \mathbf{v}}\rVert^2_F\;, \label{eq-opt_problem1}
\end{align}
\end{small}
where $\rho_1$ and $\rho_2$ are penalty parameters (we set  the $\rho = \rho_1=\rho_2$ for simpler solutions). 
By employing variable splitting algorithms such as half-quadratic splitting (HQS)~\cite{hqs}, the optimization
problem (\ref{eq-opt_problem1}) can be addressed by iteratively solving three sub-problems\footnote{Note that we follow the $t\rightarrow t-1$ iteration update order to preserve the order consistency between unrolling and the diffusion sampling process.} as follows:
\begin{small}
\begin{align}
   [\mathbf{x}|\mathbf{m}]_{t\!-\!1} \!& =\! \arg\min_{\mathbf{x},\mathbf{m}}\psi\mathcal{R}{([\mathbf{x}|\mathbf{m}])} + \frac{\rho}{2}\lVert{[\mathbf{x}|\mathbf{m}] \!- \![\mathbf{z}_{t}}|\mathbf{v}_{t}]\rVert^2_F \;, \label{eq-sub1}\\
     \mathbf{z}_{t-1}& = \arg\min_{\mathbf{z}} \frac{1}{2} \lVert{\mathbf{h} \cdot \mathbf{z}\! -\!\mathbf{y}}\rVert^2_F + \frac{\rho}{2}\lVert{\mathbf{x}_{t-1} \! -\! \mathbf{z}}\rVert^2_F \;,\label{eq-sub2} \\
       \mathbf{v}_{t-1} &= \arg\min_{\mathbf{v}}\chong{\frac{\phi}{2}}\lVert{\mathbf{v}-\mathbf{\tilde{m}}}\rVert^2_F + \frac{\rho}{2}\lVert{\mathbf{m}_{t-1} - \mathbf{v}}\rVert^2_F \;.\label{eq-sub3}
\end{align}
\end{small}
Here (\ref{eq-sub2}) and (\ref{eq-sub3}) are least-squares problems with quadratic penalty terms, which have closed-form solutions
\begin{small}
\begin{align}
\setlength\abovedisplayskip{3pt}%
\setlength\belowdisplayskip{3pt}
\mathbf{z}_{t-1} &= ( \mathbf{h} \cdot \mathbf{y} +  \rho \mathbf{x}\chong{_{t-1}} )/(\mathbf{h} \cdot \mathbf{h}+\rho \mathbf{1})\;, \label{eq-solver_sub1} \\ 
\mathbf{v}_{t-1} &= ( \chong{\phi} \mathbf{\tilde{m}} + \chong{\rho} \mathbf{m}\chong{_{t-1}}) / (\chong{\phi} + \rho)\;. \label{eq-solver_sub2}
\end{align}
\end{small}
Note that, \eqref{eq-solver_sub1} and \eqref{eq-solver_sub2} commonly refer to the data consistency (DC) steps~\cite{8067520} by sharing the information between the input and reconstructed variable. The update process of $\mathbf{x}$ and $\mathbf{m}$ can be solved by the sampling process of DMDM, denoted as $\mathcal{G}_\theta(\cdot)$ (details refer to Section~\ref{sec:dmdm}), yielding the iterates
\begin{small}
\begin{align}
[\mathbf{x}|\mathbf{m}]_{t-1} = \mathcal{G}_\theta(\mathbf{z}_t, \chong{\mathbf{y}, \mathbf{v}_t}, t)\;.
\end{align}
\end{small}
Algorithm~\ref{algo} summarizes the whole process of diffusion-based unrolling, where the $\mathcal{G}_\theta(\cdot)$ corresponds to Lines 2-3. 
The sampling of $\mathcal{G}_\theta$ follows the diffusive sampling strategy of DDIM~\cite{song2020denoising,ozdenizci2022restoring} to accelerate the inference stage.

\wh{It is noted that, in} contrast to repeatly forwarding the single-stage model, diffusion model is a natural architecture to solve the unrolling optimization problem via the progressive generation process. 
\textbf{\textit{Our framework can incorporate the degradation priors into the diffusion model with almost no additional inference time.}}
The shadow-free image is slowly restored based on the diffusion model, while the extra degradation prior can largely accelerate the shadow-free image generation and 
make the iterations close to the truly shadow-free data manifold.
The effectiveness of our design is manifested in Section~\ref{sec:ablation}.

\begin{table*}[!t]
\centering
\footnotesize
\setlength{\tabcolsep}{0.6em}
\vspace{-0.1cm}
\renewcommand{\arraystretch}{0.7}
\adjustbox{width=.78\linewidth}{
    \begin{tabular}{c|l|ccc| ccc| ccc}
        \toprule
               \multirow{2}{*}{} & \multirow{2}{*}{Method} & \multicolumn{3}{c|}{Shadow Region (S)}  &
                 \multicolumn{3}{c|}{Non-Shadow Region (NS)}  &
                 \multicolumn{3}{c}{All Image (ALL)} \\
                 & & PSNR$\uparrow$ & SSIM$\uparrow$ & RMSE$\downarrow$ & PSNR$\uparrow$ & SSIM$\uparrow$ & RMSE$\downarrow$ & PSNR$\uparrow$ & SSIM$\uparrow$ & RMSE$\downarrow$ \\
                 \midrule
                \multirow{12}{*}{\rotatebox{90}{ISTD}} & Input Image   &22.40 & 0.936 & 32.10 & 27.32 & 0.976 & 7.09 & 20.56 & 0.893 & 10.88\\
              & Guo \etal~\cite{guo2012paired}  & 27.76 & 0.964 & 18.65 & 26.44 & 0.975 & 7.76 & 23.08 & 0.919 & 9.26 \\
              &MaskShadow-GAN~\cite{hu2019mask} & - & - & 12.67 & - & - & 6.68 & - & -& 7.41\\
                &ST-CGAN~\cite{wang2018stacked} & 33.74 & 0.981 & 9.99 & 29.51 & 0.958 & 6.05 & 27.44 & 0.929 & 6.65\\
                    &DSC~\cite{hu2019direction} & 34.64 & 0.984 & 8.72 & 31.26 & 0.969 & 5.04 & 29.00 & 0.944 & 5.59\\
                    & G2R~\cite{liu2021from} & 31.63 & 0.975 & 10.72 & 26.19 & 0.967 & 7.55 & 24.72 & 0.932 & 7.85\\
                       & DHAN~\cite{cun2020towards} & 35.53 & 0.988 & 7.49 & 31.05 & 0.971 & 5.30 & 29.11 & 0.954 & 5.66\\
                     & Fu~\etal~\cite{fu2021auto} & 34.71 & 0.975 & 7.91 & 28.61 & 0.880 & 5.51 & 27.19 & 0.945 & 5.88\\
                       & DC-ShadowNet~\cite{jin2021dc} & 31.69 & 0.976 & 11.43 & 28.99 & 0.958 & 5.81 & 26.38 & 0.922 & 6.57\\
                       & Zhu~\etal~\cite{zhu2022efficient} & 36.95 & 0.987 & 8.29 & 31.54&0.978 & 4.55 & 29.85 & 0.960 & 5.09\\
            & BMNet~\cite{zhu2022bijective} & 35.61 & 0.988 & 7.60 & 32.80 & 0.976 & 4.59 & 30.28 & 0.959 & 5.02 \\
                       &     \cellcolor{Gray}Ours &\cellcolor{Gray}\textbf{40.15} & \cellcolor{Gray}\textbf{0.994} & \cellcolor{Gray}\textbf{4.13}&  \cellcolor{Gray}\textbf{33.70} & \cellcolor{Gray}\textbf{0.977} & \cellcolor{Gray}\textbf{4.14} &\cellcolor{Gray}\textbf{32.33} &\cellcolor{Gray}\textbf{0.969} &\cellcolor{Gray}\textbf{4.12}  \\
\midrule
\multirow{10}{*}{\rotatebox{90}{SRD}} & Input Image   &18.96 & 0.871 & 36.69 & 31.47 & 0.975 & 4.83 & 18.19 & 0.830 & 14.05\\
              & Guo \etal~\cite{guo2012paired}  & - & - & 29.89 & - & - & 6.47 & - & - & 12.60 \\
                &DeshadowNet~\cite{qu2017deshadownet}& - &-&  11.78 & -&- & 4.84&-&- & 6.64 \\
               & DSC~\cite{hu2019direction} & 30.65 & 0.960 & 8.62 & 31.94 & 0.965 & 4.41 & 27.76 & 0.903 & 5.71\\
               & DHAN~\cite{cun2020towards} & 33.67 & 0.978 & 8.94 & 34.79 & 0.979 & 4.80 & 30.51 & 0.949 & 5.67\\
               & Fu~\etal~\cite{fu2021auto} & 32.26 & 0.966 & 9.55 & 31.87 & 0.945 & 5.74 & 28.40 & 0.893 & 6.50 \\
                & DC-ShadowNet~\cite{jin2021dc} & 34.00 & 0.975 & 7.70 & 35.53 & 0.981 & 3.65 & 31.53 & 0.955 & 4.65\\
                 &    Zhu~\etal~\cite{zhu2022efficient} & 34.94 & 0.980 & 7.44 & 35.85 & 0.982 & 3.74 &31.72 & 0.952 & 4.79\\
                &BMNet~\cite{zhu2022bijective}& 35.05 & 0.981 & 6.61 & 36.02 & 0.982 & 3.61 & 31.69 & 0.956 & 4.46 \\
                      &      \cellcolor{Gray}Ours &\cellcolor{Gray}\textbf{38.72} & \cellcolor{Gray}\textbf{0.987} & \cellcolor{Gray}\textbf{4.98}&  \cellcolor{Gray}\textbf{37.78} & \cellcolor{Gray}\textbf{0.985} & \cellcolor{Gray}\textbf{3.44} &\cellcolor{Gray}\textbf{34.73} &\cellcolor{Gray}\textbf{0.970} &\cellcolor{Gray}\textbf{3.63}  \\
\bottomrule
    \end{tabular}
}
\vspace{-0.2cm}
\caption{The quantitative results of shadow removal
using our ShadowDiffusion and recent methods on ISTD~\cite{wang2018stacked} and SRD~\cite{qu2017deshadownet}
datasets.}
\vspace{-0.2cm}
\label{tab:srd_res}
\end{table*}

\begin{table}[!t]
\centering
\footnotesize
\setlength{\tabcolsep}{0.3em}
\renewcommand{\arraystretch}{0.7}
\adjustbox{width=1.\linewidth}{
    \begin{tabular}{l|cc|cc|cc}
        \toprule
         \multirow{2}{*}{Method} & \multicolumn{2}{c|}{Shadow}  & \multicolumn{2}{c|}{Non-Shadow}& \multicolumn{2}{c}{All}\\
         & PSNR$\uparrow$ & RMSE$\downarrow$  &  PSNR$\uparrow$ & RMSE$\downarrow$  &  PSNR$\uparrow$ & RMSE$\downarrow$ \\
                 \midrule
                 Input Image  & 20.83 & 40.2 & 37.46 & 2.6 &20.46 & 8.5\\
               
                DeshadowNet~\cite{qu2017deshadownet} & -& 15.9 & - & 6.0 & -& 7.6 \\
                ST-CGAN~\cite{wang2018stacked} & -& 13.4 & -& 7.7 & -& 8.7\\
                Param-Net~\cite{le2020shadow} & - &9.7 &-& 3.0 &-& 4.0\\
                SP+M-Net~\cite{le2019shadow} &37.59& 5.9 & 36.02 & 3.0 & 32.94 & 3.5\\
                 DHAN~\cite{cun2020towards} & 32.92 & 11.2 & 27.15 & 7.1 & 25.66 & 7.8\\
                Fu~\etal~\cite{fu2021auto} & 36.04 & 6.6 & 31.16 & 3.8 &29.45& 4.2 \\
                BMNet~\cite{zhu2022bijective}& - &5.6 & - & 2.5 & -& 3.0 \\
                            \cellcolor{Gray}Ours &\cellcolor{Gray}\textbf{39.82} &\cellcolor{Gray}\textbf{4.9} &\cellcolor{Gray}\textbf{38.90}& \cellcolor{Gray}\textbf{2.3} & \cellcolor{Gray}\textbf{35.72} & \cellcolor{Gray}\textbf{2.7}\\
\bottomrule
    \end{tabular}
}
\vspace{-0.2cm}
\caption{The quantitative results of shadow removal
using our ShadowDiffusion and recent methods on ISTD+~\cite{le2019shadow}
dataset.}
\vspace{-0.2cm}
\label{tab:istdplus_res}
\end{table}

\vspace{-5pt}
\section{Experiments}
\vspace{-2pt}
\subsection{Experimental Setups}
\vspace{-3pt}
\noindent\textbf{Implementation details.}
The proposed method is implemented using PyTorch, which is trained using one NVIDIA RTX A5000 GPU. 
The training epoch is set as 1000. We use Adam optimizer with the momentum as $(0.9,0.999)$. The initial learning rate is $3\times 10^{-5}$.
Following~\cite{saharia2022image}, we use the Kaiming initialization technique~\cite{he2015delving} to initialize the weights of the proposed model and use $0.9999$ Exponential Moving Average (EMA) for all our experiments. We followed the similar U-Net architecture as denoiser $\bm{\epsilon}_\theta$ of~\cite{saharia2022image}. We used $1000$ diffusion steps $T$ and noise schedule $\beta_t$ linearly increasing from $0.0001$ to $0.02$ for training, and $25$ steps for inference.
We select the most recent transformer-based image-to-image backbone~\cite{wang2022uformer} as the degradation estimation network $\mathcal{D}$, in which we pre-train $\mathcal{D}$ with the concatenation of shadow image and mask as input and regard the $\mathbf{h}_{gt}=\mathbf{y}/(\mathbf{x}+\eta) $, with $\eta=1\times e^{-4}$, as the ground truth degradation map.
We set the $\lambda=0.5$ in our experiments.
The detailed architectures and \wh{hyper-parameter} settings can be found in the \textbf{supplementary}.

\noindent\textbf{Benchmark datasets.} 
We work with three benchmark datasets for the various shadow removal experiments: 
(1) ISTD~\cite{wang2018stacked} dataset includes 1330 training and 540 testing triplets (shadow images, masks and shadow-free images).
(2) Adjusted ISTD (ISTD+) dataset~\cite{le2019shadow} reduces the illumination inconsistency between the shadow and shadow-free image of ISTD. 
(3) SRD~\cite{qu2017deshadownet} dataset consists of 2680 training and 408 testing pairs of shadow and shadow-free images. 
We use the predicted masks that are provided by DHAN~\cite{cun2020towards} for training and testing following most previous methods~\cite{fu2021auto,cun2020towards,zhu2022efficient,zhu2022bijective}.

\noindent\textbf{Evaluation measures.} Following the previous works~\cite{wang2018stacked,guo2012paired,qu2017deshadownet,le2019shadow,cun2020towards,fu2021auto}, we utilize the root mean square error (RMSE) in the LAB color space as the \textbf{quantitative} evaluation metric of the shadow removal results, comparing to the ground truth shadow-free images. 
Besides, we also adopt the Peak Signal-to-Noise Ratio (PSNR) and the structural similarity (SSIM)~\cite{wang2004image} to measure the performance of various methods in the RGB color space. For the PSNR and SSIM metrics, higher values represent better results.

\subsection{Comparison with State-of-the-Art}
We compare the proposed method with the popular or state-of-the-art shadow removal algorithms, including one traditional method, \ie, Guo~\etal~\cite{guo2012paired}, and several deep learning-based methods, \ie, MaskShadow-GAN~\cite{hu2019mask}, DeshadowNet~\cite{qu2017deshadownet}, ST-CGAN~\cite{wang2018stacked}, DSC~\cite{hu2019direction}, 
G2R~\cite{liu2021from}, DHAN~\cite{cun2020towards},
 Param-Net~\cite{le2020shadow},
 SP+M-Net~\cite{le2019shadow}, Fu~\etal~\cite{fu2021auto}, DC-ShadowNet~\cite{jin2021dc}, Zhu~\etal~\cite{zhu2022efficient}, and BMNet~\cite{zhu2022bijective}.
All of the shadow removal results by the competing methods are quoted from the original papers or reproduced using their official implementations.
Following the previous methods~\cite{fu2021auto,zhu2022efficient,zhu2022bijective} to evaluate the shadow removal performance,
we evaluate the shadow \wh{removal} results with a resolution
of $256\times 256$.

\begin{figure*}[!t]
\centering
\vspace{-1mm}
\includegraphics[width=.95\linewidth]{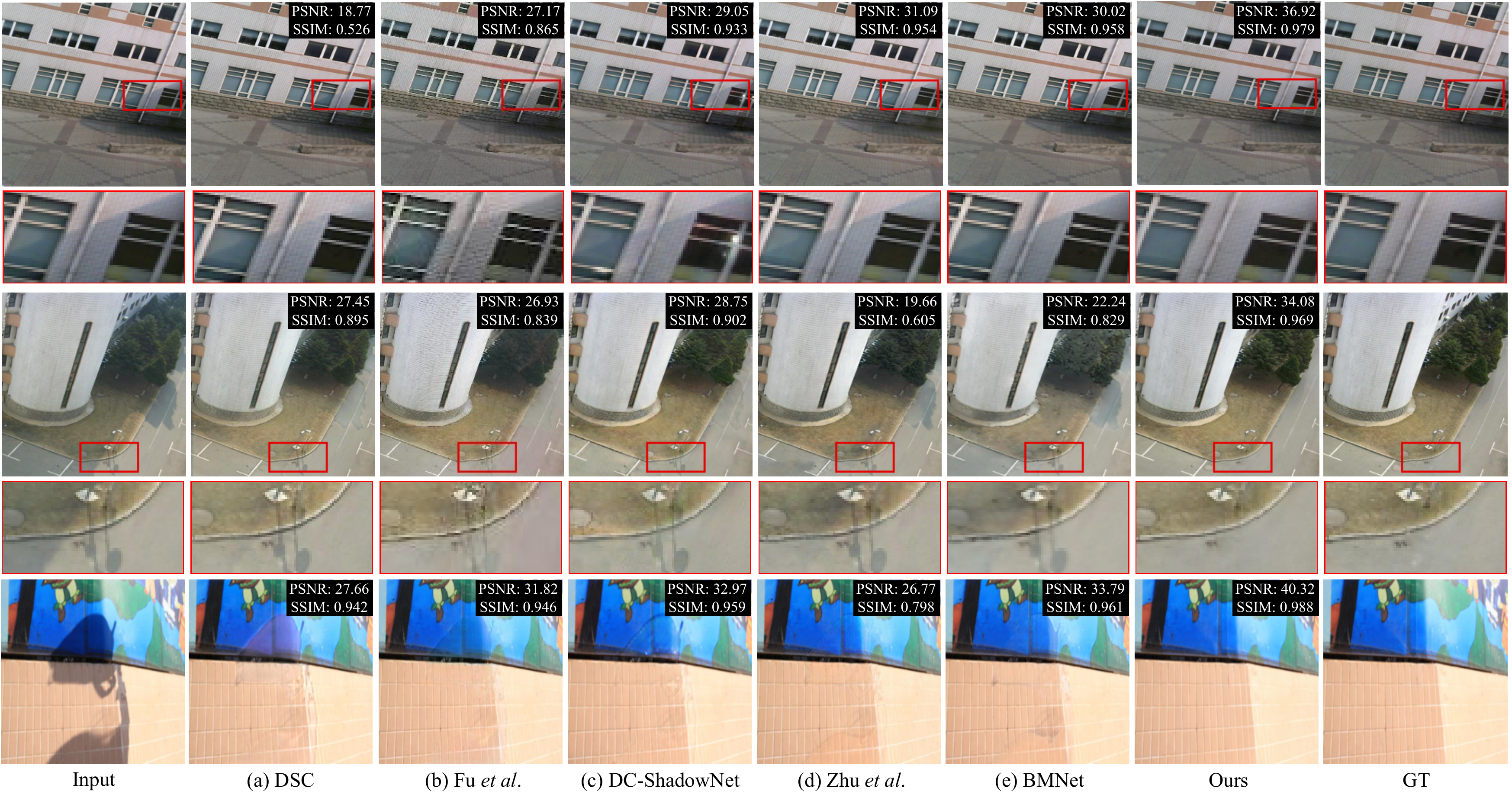} 
\vspace{-2mm}
\caption{Examples of shadow removal results on the SRD~\cite{qu2017deshadownet} dataset. The input shadow image, the estimated results of (a) DSC~\protect\cite{hu2019direction}, (b) Fu~\etal~\protect\cite{fu2021auto}, (c) DC-ShadowNet~\protect\cite{jin2021dc}, (d) Zhu~\etal~\cite{zhu2022efficient}, (e) BMNet~\cite{zhu2022bijective}, and Ours, as well as the ground truth, respectively. Please \textbf{zoom in} to see the details.}
\vspace{-0.2cm}
\label{fig:srd_res} 
\end{figure*}

\noindent\textbf{Quantitative evaluation.}
Tables~\ref{tab:srd_res}\&\ref{tab:istdplus_res} show the quantitative results on the testing sets over ISTD, SRD, and ISTD+, respectively.
It is clear that our methods outperform all competing methods by large margins in \wh{the} shadow area, non-shadow area and the whole image over all of the three datasets.
It significantly improves the PSNR from 31.69dB to 34.73dB over SRD dataset, compared to the most recent method BMNet, especially for the shadow region from 35.05dB to 38.72dB.
The shadow scenarios from SRD dataset are much more complicated than ISTD dataset, even including some object-correlated shadows, \eg, the shadows are caused by the trees and building in the second row of Figure~\ref{fig:srd_res}. 
Existing methods may ignore the effective degradation and generative priors, failing in such complicated textures.
Conversely, 
with the \wh{merits} of the joint image-mask modeling, our methods can better deal with \wh{object-correlated} shadow cases.
Besides, by incorporating the useful illumination degradation assumption and learning the desired shadow-free image distribution, our method can handle \wh{complicated} cases and produce artifact-free results.

\begin{figure}[!t]
\centering
\vspace{-2mm}
\includegraphics[width=1.\linewidth]{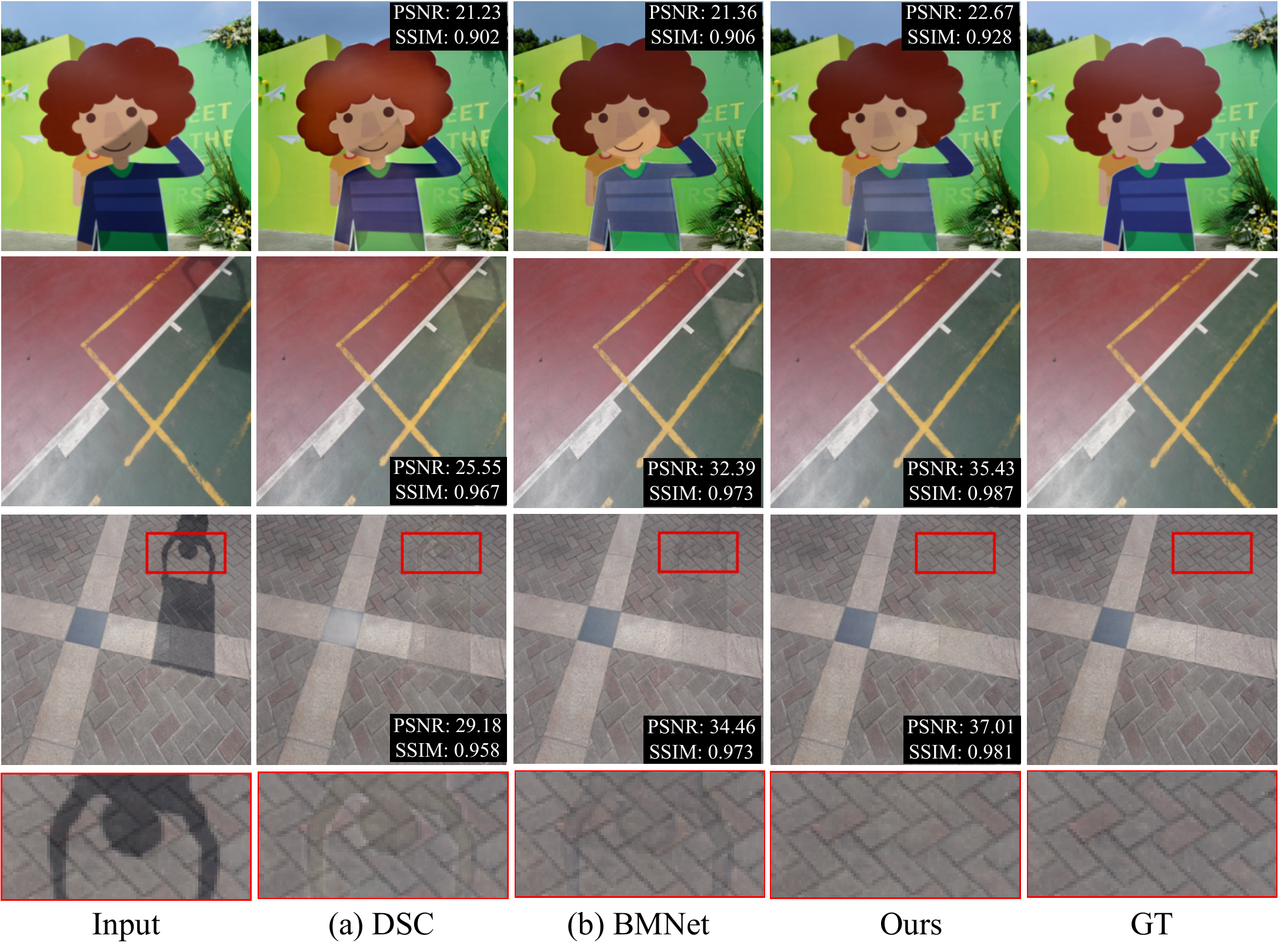} 
\vspace{-6mm}
\caption{Examples of shadow removal results on the ISTD~\cite{wang2018stacked} dataset. The input shadow image, the estimated results of (a) DSC~\protect\cite{hu2019direction}, (b) BMNet~\cite{zhu2022bijective}, and Ours, as well as the ground truth, respectively. Please \textbf{zoom in} to see the details.}
\vspace{-0.2cm}
\label{fig:istd_res} 
\end{figure}

\begin{figure}[!t]
\centering
\vspace{-2mm}
\includegraphics[width=.9\linewidth]{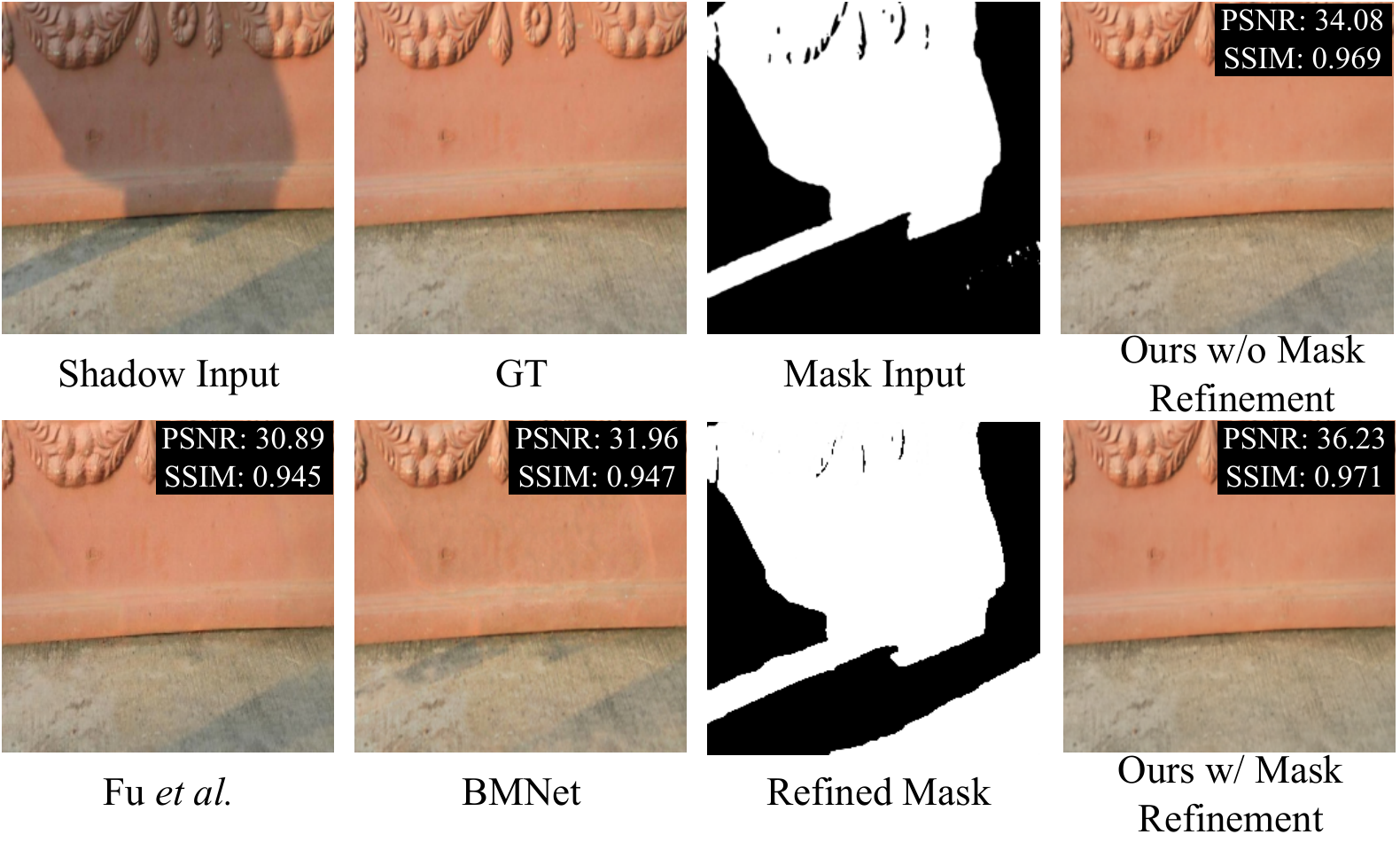} 
\vspace{-4mm}
\caption{Visual examples of the results of our model w/ and w/o mask refinement, as well as two recent competing methods, \ie, Fu~\etal~\cite{fu2021auto} and BMNet~\cite{zhu2022bijective}.}
\vspace{-0.2cm}
\label{fig:mask_refine} 
\end{figure}

\noindent\textbf{Qualitative evaluation.}
To further demonstrate the advantage of our method against other competing methods, 
Figures~\ref{fig:srd_res} \& \ref{fig:istd_res} present the visual examples of the shadow removal results on SRD and ISTD datasets, respectively.
More visual examples can be found in the \textbf{supplementary}.
Note that the images from the SRD dataset have more complicated textures and color distributions.
In these samples of the SRD dataset, previous works fail to enhance the illumination of the background and suppress the boundary artifacts in a complicated and colorful region, \eg, the blue poster of the third example in Figure~\ref{fig:srd_res}.
Almost all competing methods cannot preserve the illumination consistencies between shadow and non-shadow regions, which seriously \wh{destroys} the image structures and patterns as shown in the wall of the first and third examples in Figure~\ref{fig:srd_res}.
Some small shadow regions, \eg, the shadow caused by the street lamp of the second example of Figure~\ref{fig:srd_res}, are easily ignored by existing methods.
Instead, thanks to the auxiliary of the mask refinement, the shadow mask of fine areas can be more accurate, achieving  \wh{better} results for different sizes of shadows.
On the other hand, the image from ISTD dataset \wh{has} high context similarity and the scene is relatively simple, while the shadow residual would be more visible.
In these samples of the ISTD dataset, previous works usually produce illumination inconsistencies and wrongly-enhanced shadow boundaries.
The DSC~\cite{hu2019direction} would wrongly enlighten some regions with insufficient lightness, \eg, the black floor tile in \wh{the} third example in Figure~\ref{fig:istd_res}, leading to many ghosts.
Moreover, due to the limited dataset, methods without effective prior always have poor scene \wh{understanding} and generalizability.
The performance would largely degrade when generalizing to unusual cases, \eg, the first example in Figure~\ref{fig:istd_res}, where images with stronger contrast and face patterns are scarce in the training set.
However, with the merits of employed shadow degradation and generative priors, it is clear that our methods can successfully produce natural shadow-free images without boundary artifacts and shadow patterns.

\vspace{-2pt}
\subsection{Ablation Study}\label{sec:ablation}
\vspace{-3pt}
\noindent\textbf{The effect of iterative mask refinement.}
Previous methods~\cite{chen2021canet,cun2020towards,fu2021auto,zhu2022bijective} employed the detected shadow masks as additional auxiliary information of the network to provide the shadow locations.
However, the detected masks might contain inaccurate regions or coarse boundaries as shown in Figure~\ref{fig:mask_refine}. Such wrong guidance makes almost all mask-guided shadow removal algorithms fail, especially for the SRD dataset~\cite{qu2017deshadownet}, since the complicated scenes and diverse shadow shapes.
With the merits of the proposed dynamic mask-aware diffusion model (DMDM), the coarse or inaccurate mask can be iteratively corrected along with the shadow-free image generation.
Thus, the \wh{effect} of the wrong mask can be effectively alleviated and our model has better robustness in practical applications.
Table~\ref{tab:mask_refine} shows that the shadow removal results of our model with and without mask refinement under the guidance of masks generated by different shadow detectors~\cite{cun2020towards,zhu2021mitigating} over SRD dataset.
The shadow removal results will be better if the initial mask is more accurate as shown in Table~\ref{tab:mask_refine}, and the decline with a worse mask input will not be too noticeable with the proposed iterative mask refinement.

\begin{table}[!t]
\centering
\footnotesize
\renewcommand{\arraystretch}{0.7}
\adjustbox{width=.8\linewidth}{
    \begin{tabular}{l|cc|cc}
        \toprule
         \multirow{2}{*}{Method} & \multicolumn{2}{c|}{Detector1~\cite{cun2020towards}}  & \multicolumn{2}{c}{Detector2~\cite{zhu2021mitigating}}\\
         & PSNR$\uparrow$ & SSIM$\uparrow$  &  PSNR$\uparrow$ & SSIM$\uparrow$   \\
                 \midrule
               
                w/o mask refine & 34.55 & 0.968 & 33.61& 0.943 \\
              \rowcolor{Gray}  w/ mask refine & \textbf{34.73} & \textbf{0.970} & \textbf{34.27}& \textbf{0.952}\\
\bottomrule
    \end{tabular}
}
\vspace{-0.2cm}
\caption{Quantitative comparison between results produced by w/ and w/o mask refinement on SRD~\cite{qu2017deshadownet}
dataset using the detected shadow mask from different shadow detectors as initial mask $\tilde{m}$.}
\label{tab:mask_refine}
\end{table}

\begin{table}[!t]
\centering
\footnotesize
\setlength{\tabcolsep}{0.4em}
\renewcommand{\arraystretch}{0.7}
\adjustbox{width=1.\linewidth}{
    \begin{tabular}{l|cc|cc|cc}
        \toprule
         \multirow{2}{*}{Method} & \multicolumn{2}{c|}{ISTD}  & \multicolumn{2}{c|}{ISTD+}& \multicolumn{2}{c}{SRD}\\
         & PSNR$\uparrow$ & SSIM$\uparrow$  &  PSNR$\uparrow$ & SSIM$\uparrow$  &  PSNR$\uparrow$ & SSIM$\uparrow$ \\
                 \midrule
               
               Ours w/o unrolling & 32.12 & 0.964 &35.21 &0.967 & 34.36& \textbf{0.970}\\
              \rowcolor{Gray} Ours (Complete model) & \textbf{32.33} & \textbf{0.969} & \textbf{35.72} & \textbf{0.969} & \textbf{34.73} & \textbf{0.970}\\
\bottomrule
    \end{tabular}
}
\vspace{-0.2cm}
\caption{Quantitative comparison between the results produced by w/o unrolling, and the complete model over ISTD~\cite{wang2018stacked}, ISTD+~\cite{le2019shadow}, and SRD~\cite{qu2017deshadownet} datasets.}
\label{tab:unrolling}
\end{table}

\begin{figure}[!t]
\centering
	\begin{center}
		\begin{tabular}{c@{ }c}
			\includegraphics[width=.35\linewidth]{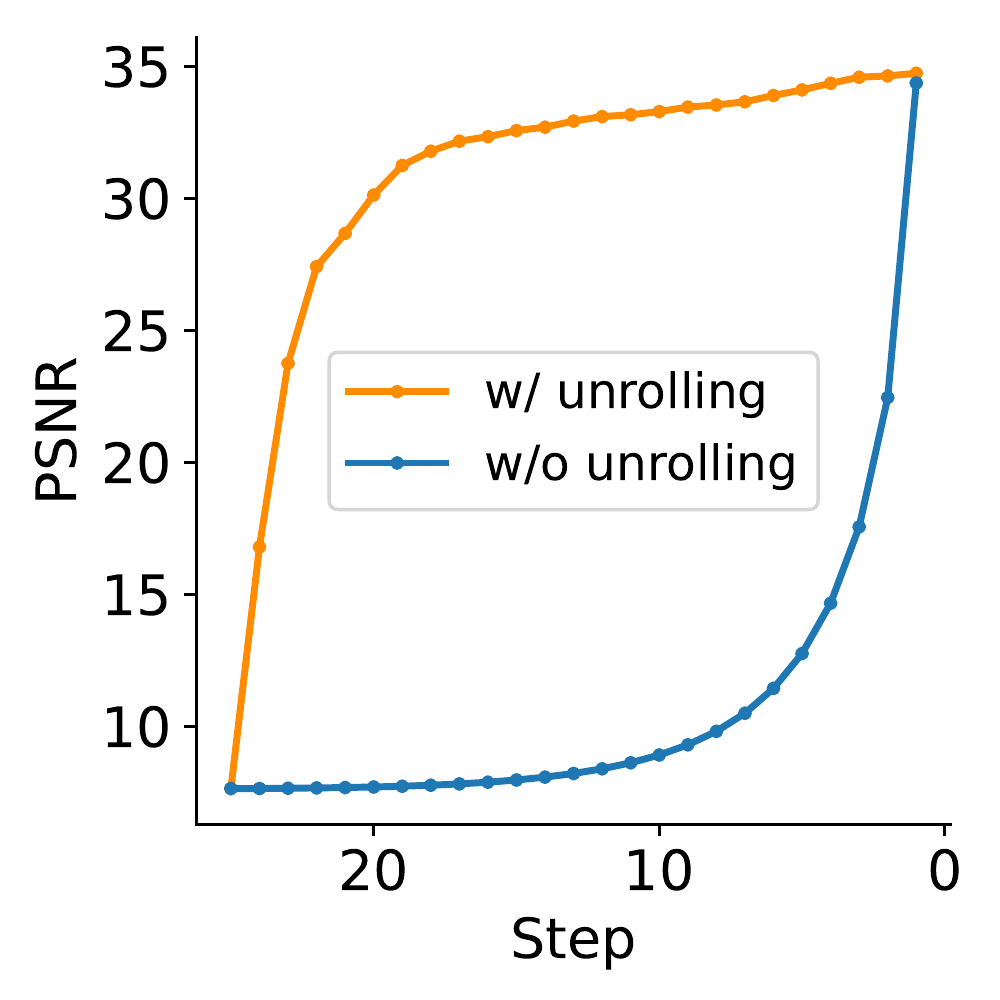}~&
			\includegraphics[width=.55\linewidth]{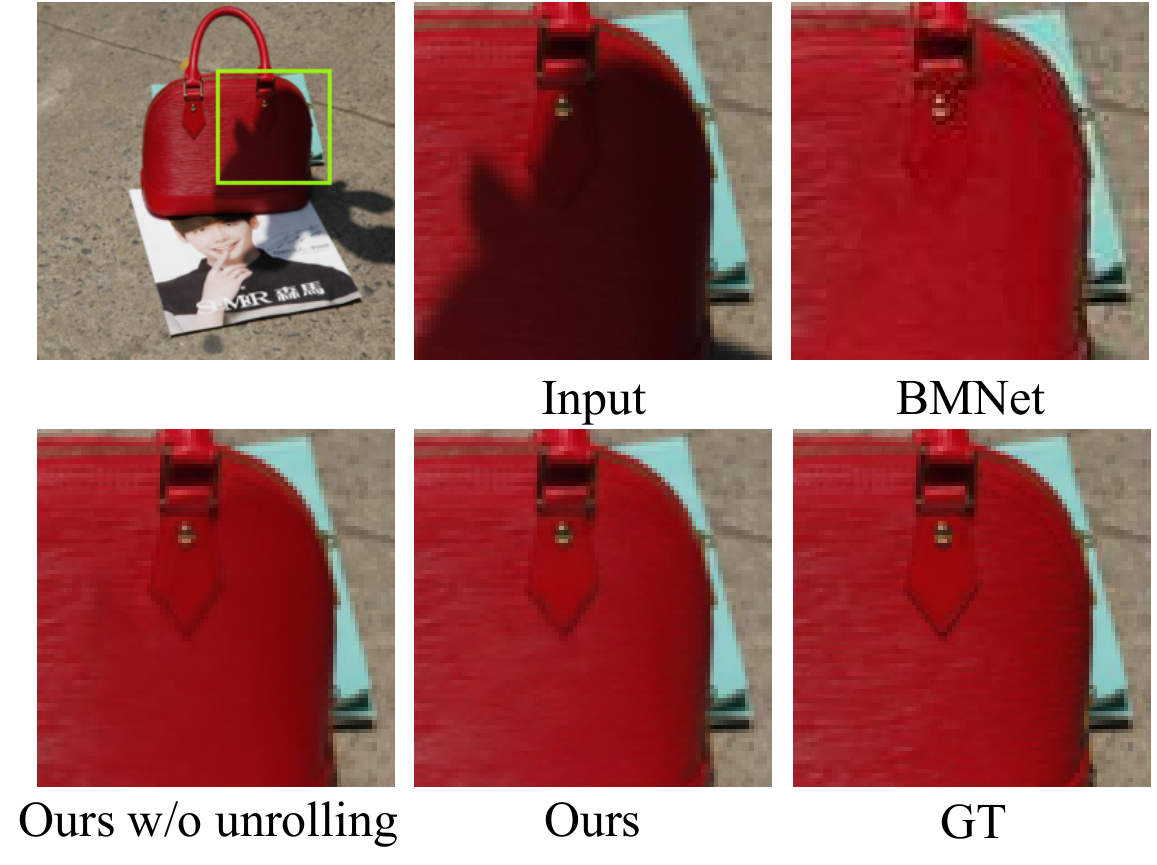}\\
		\end{tabular}
	\end{center}
\vspace{-6mm}
\caption{Left: The PSNR performance variations of our model w/o and w/ unrolling with steps. Right: Visual examples of the results of the most recent competing method, \ie, BMNet~\cite{zhu2022bijective}, Our model w/o unrolling, and Our Complete model.}
\vspace{-0.2cm}
\label{fig:iter} 
\end{figure}

\noindent\textbf{The effect of unrolling framework.}
In order to illustrate the effectiveness of our unrolling framework,
we first investigate the performance of only our dynamic mask-aware diffusion model (DMDM) without the unrolling optimization as shown in the first row in Table~\ref{tab:unrolling}. 
We observe that the performance of \wh{the} complete model is better than the separate diffusion model over all testing sets.
The unrolling optimization can provide \wh{reliable} guidance in the sampling stage of \wh{the} diffusion model, producing a more accurate exposure and artifact-free results than the model without unrolling as shown in Figure~\ref{fig:iter} (right).
However, according to a predicted shadow degradation map, the diffusion model can be boosted in the sampling stage. Besides, the performance of \wh{the} diffusion model only would be converged very slow in the sampling stage and the output would be unstable even around the last three steps as shown in Figure~\ref{fig:iter} (left).
Under the strict guidance of degradation map, the convergence would be faster and more stable.

\noindent\textbf{The effect of different diffusion models.}
We select two recent diffusion-based image restoration methods, \ie, SR3~\cite{saharia2022image} and WeatherDiffusion~\cite{ozdenizci2022restoring} as the competing methods to verify the effectiveness of the proposed dynamic mask-aware diffusion model (DMDM) and the proposed diffusion-based unrolling framework as shown in Table~\ref{tab:diffusion}. 
For a fair evaluation, we re-train these two methods and change the original three-channel condition (shadow image) into four-channel (the concatenation of shadow image and mask).
Obviously, the existing diffusion model performs much worse for shadow removal than our proposed DMDM.
Only the concatenation of shadow image and mask as \wh{a} condition cannot provide a sufficient prior for shadow-free image generation, especially for real-world shadow removal with limited training pairs, comparing the first and fourth rows in Table~\ref{tab:diffusion}.

\subsection{Extension to Other Image Enhancement Tasks}
\lanqing{
Our ShadowDiffusion can be easily \wh{applied} to other image enhancement tasks, \eg, low-light enhancement and exposure correction, whose degradation can be regarded as the special case of proposed degradation Model \eqref{eq:shadow_model} and the $m$ will be the all $1$ matrix.
Note that we remove the mask refinement and mask conditions in the framework since other enhancement tasks are globally \wh{corrupted} without the provided mask.
Our \wh{revised} ShadowDiffusion also achieves new state-of-the-art performances among these two tasks as shown in Table~\ref{tab:other}.
More comparison results can be found in the \textbf{supplementary}.}

\begin{table}[!t]
\centering
\footnotesize
\setlength{\tabcolsep}{0.4em}
\renewcommand{\arraystretch}{0.7}
\adjustbox{width=1.\linewidth}{
    \begin{tabular}{l|cc|cc|cc}
        \toprule
         \multirow{2}{*}{Method} & \multicolumn{2}{c|}{Shadow}  & \multicolumn{2}{c|}{Non-Shadow}& \multicolumn{2}{c}{All}\\
         & PSNR$\uparrow$ & SSIM$\uparrow$  &  PSNR$\uparrow$ & SSIM$\uparrow$  &  PSNR$\uparrow$ & SSIM$\uparrow$ \\
                 \midrule
               
                SR3~\cite{saharia2022image} & 35.44 & 0.980 & 34.35& 0.970 & 31.29 & 0.946\\
                WeatherDiffusion~\cite{ozdenizci2022restoring} & 33.38 & 0.981 & 31.15& 0.972 & 28.45& 0.951\\
                \midrule
                DMDM Only & 38.39& \textbf{0.987}  & 37.21& 0.982 &34.36 & \textbf{0.970} \\
               \rowcolor{Gray} Ours (Complete model) & \textbf{38.72} & \textbf{0.987} & \textbf{37.78} & \textbf{0.985} & \textbf{34.73} & \textbf{0.970} \\

\bottomrule
    \end{tabular}
}
\vspace{-0.2cm}
\caption{Quantitative comparisons with different diffusion-based models over SRD dataset~\cite{qu2017deshadownet}.}
\label{tab:diffusion}
\end{table}

\begin{table}[!t]
\centering
\footnotesize
\setlength{\tabcolsep}{0.3em}
\renewcommand{\arraystretch}{0.7}
\adjustbox{width=.95\linewidth}{
    \begin{tabular}{l|ccc|l|cc}
        \toprule
         \multicolumn{4}{c|}{{Low-light enhancement}} & \multicolumn{3}{c}{{Exposure correction}} \\
         Method
         & PSNR$\uparrow$ & SSIM$\uparrow$  &  LPIPS$\downarrow$  &  Method & PSNR$\uparrow$ & SSIM$\uparrow$ \\
                 \midrule
                 KinD++~\cite{zhang2021beyond} & 21.30 &0.82 &0.16 & Deep UPE~\cite{wang2019underexposed} & 14.25& 0.64 \\
     URetinex-Net~\cite{wu2022uretinex} & 21.33 &0.83 &0.12 & DPE (HDR)~\cite{chen2018deep}&  16.21 &0.62 \\
     MIRNet~\cite{Zamir2020MIRNet} & {24.14} & {0.84} & 0.13 & 
     Afifi~\etal~\cite{afifi2021learning} &19.48& 0.74\\
              \rowcolor{Gray} Ours & {\textbf{27.36}} & {\textbf{0.93}} & {\textbf{0.10}} & Ours &  \textbf{22.33} & \textbf{0.84}\\
\bottomrule
    \end{tabular}
}
\vspace{-0.2cm}
\caption{A comparison of the recent state-of-the-art methods for (left) low-light enhancement and (right) exposure correction. }
\label{tab:other}
\end{table}

\vspace{-2pt}
\section{Conclusion}
\vspace{-2pt}
In this paper, we propose a spatially-variant shadow degradation model, decomposing the shadow degradation map into the shadow mask and shadow intensity. Inspired by that, we propose an unrolling diffusion framework, dubbed as ShadowDiffusion, to explicitly integrate degradation prior and diffusive generative prior.
Moreover, we further consider \wh{mask} refinement as an auxiliary task of the diffusion generator to progressively refine the shadow mask. 
Finally, comprehensive experiments \wh{demonstrate} the superiority of our ShadowDiffusion, which achieves significant improvement compared to the state-of-the-art methods over ISTD, ISTD+, and SRD datasets.

{\small
\bibliographystyle{ieee_fullname}
\bibliography{egbib}
}

\end{document}